\documentclass[pdflatex,sn-mathphys-num]{sn-jnl}

\usepackage{graphicx}%

\usepackage{geometry}

\AtBeginDocument{%
    \newgeometry{
        left=1.6cm,
        right=1.6cm,
        top=1.7cm,
        bottom=1.7cm
    }
    \renewcommand{\figurename}{Figure}
}

\usepackage{multirow}%
\usepackage{amsmath,amssymb,amsfonts}%
\usepackage{amsthm}%
\usepackage{mathrsfs}%
\usepackage[title]{appendix}%
\usepackage{xcolor}%
\usepackage{textcomp}%
\usepackage{manyfoot}%
\usepackage[inkscapelatex=false]{svg}
\usepackage{booktabs}%
\usepackage{algorithm}%
\usepackage{algorithmicx}%
\usepackage{algpseudocode}%
\usepackage{listings}%
\usepackage{xcolor}
\usepackage{cleveref}
\usepackage{siunitx}
\crefname{figure}{Figure}{Figures}
\Crefname{figure}{Figure}{Figures}

\crefname{table}{Table}{Tables}
\Crefname{table}{Table}{Tables}

\crefname{section}{Section}{Sections}
\Crefname{section}{Section}{Sections}

\crefname{subsection}{Section}{Sections}
\Crefname{subsection}{Section}{Sections}

\crefname{subsubsection}{Section}{Sections}
\Crefname{subsubsection}{Section}{Sections}

\crefname{equation}{Equation}{Equations}
\Crefname{equation}{Equation}{Equations}

\usepackage{longtable}%
\usepackage[none]{hyphenat}
\usepackage{xspace}%
\usepackage{comment}

\theoremstyle{thmstyleone}%
\theoremstyle{thmstyletwo}%

\theoremstyle{thmstylethree}%

\newcommand{\model}{Lumen\xspace}
\newcommand{\papertitle}{%
\model: Parameter-Efficient Alignment of Pretrained Vision and Language Encoders for Zero-Shot Computational Pathology%
}

\title{\papertitle}

\begin{document}




\author[1,2]{\fnm{Kiarash} \sur{Tajbakhsh}}
\equalcont{These authors contributed equally to this work.}
\author[1,2]{\fnm{Abdelrahman} \sur{Faqieh}}
\equalcont{These authors contributed equally to this work.}
\author[1,2,3]{\fnm{Michael} \sur{Jopiti}}
\equalcont{These authors contributed equally to this work.}
\author[1,2,3]{\fnm{Javier} \sur{Garcia-Baroja}}
\author[1,4]{\fnm{Philipp} \sur{Zens}}
\author[1,2]{\fnm{Branislav} \sur{Zagrapan}}
\author[5]{\fnm{Yuri} \sur{Tolkach}}
\author[6]{\fnm{Martin D.} \sur{Berger}}
\author[1,2]{\fnm{Aurel} \sur{Perren}}
\author[1,2]{\fnm{Bastian} \sur{Dislich}}
\author[1,2]{\fnm{Inti} \sur{Zlobec}}
\author*[1,2]{\fnm{Amjad} \sur{Khan}}\email{amjad.khan@unibe.ch}



\affil[1]{\orgdiv{Institute of Tissue Medicine and Pathology}, \orgname{University of Bern}, \orgaddress{\country{Switzerland}}}

\affil[2]{\orgdiv{Department of Digital Medicine},
\orgname{University of Bern}, \orgaddress{\country{Switzerland}}}
 
\affil[3]{\orgdiv{Graduate School for Cellular and Biomedical Sciences}, \orgname{University of Bern}, \orgaddress{\country{Switzerland}}}

\affil[4]{\orgdiv{Graduate School for Health Sciences}, \orgname{University of Bern}, \orgaddress{\country{Switzerland}}}


\affil[5]{\orgdiv{Institute of Pathology}, \orgname{University Hospital Cologne, Medical Faculty}, \orgaddress{\country{Germany}}}

\affil[6]{\orgdiv{Department of Medical Oncology}, \orgname{Inselspital, University Hospital of Bern}, \orgaddress{\country{Switzerland}}}


\abstract{{\unboldmath
Pathology vision--language models are commonly built by pretraining or fine-tuning large encoders on paired image--caption data. We asked whether a pathology vision–-language model can instead be assembled by parameter--efficient alignment of frozen unimodal foundation models, leaving their pretrained representations untouched. Here we present \model, which aligns frozen Virchow2 and BioMedBERT backbones using rank-4 adapters and projection heads, training only \qty{0.40}{\percent} of the total parameters on the public QUILT-1M corpus. Across nine public zero-shot patch benchmarks, \model achieved the highest mean chance-corrected balanced accuracy, $0.546$ versus $0.461$ for the strongest baseline (paired difference $0.086$, \qty{95}{\percent} CI $0.042$--$0.136$). On lymph-node metastasis detection, \model reached an AUROC of \qty{0.964}{} (\qty{95}{\percent} CI \qty{0.956}{}--\qty{0.971}{}) on \num{4214} held-out internal slides and \qty{0.955}{} (\qty{95}{\percent} CI \qty{0.942}{}--\qty{0.966}{}) on \num{2368} slides across nine external cohorts and six organs. At the internally calibrated threshold, it outperformed all vision-language baselines, with a balanced accuracy of $0.909$ (\qty{95}{\percent} CI $0.896$--$0.923$) internally and $0.915$ (\qty{95}{\percent} CI $0.902$--$0.929$) externally. \model performed competitively across the evaluations, with the exception of cross-modal retrieval, where it ranked third behind CONCH and PathGen-L/14. Fully fine-tuning both encoders gave \model no consistent benefit over low-rank adaptation, although it improved retrieval. Aligning frozen unimodal foundation models therefore yields strong and transferable performance at patch and slide level while training only a small fraction of the parameters.}}


\maketitle

\section{Introduction} \label{sec:intro}

Histopathology is central to the diagnosis and characterization of malignant disease \cite{verghese2023computational}. The transition from glass slides to digital whole-slide images (WSIs) has enabled computational analysis at scale \cite{song2023ai}. More recently, pathology foundation models trained on millions of WSIs have produced visual representations that transfer across a wide range of downstream tasks, reducing the need to train task-specific representations from scratch. However, purely visual models still require labeled examples to associate these representations with a particular diagnostic task, and collecting annotations for the breadth of questions encountered in pathology remains laborious.

Vision--language models (VLMs) relax this constraint by grounding visual representations in natural-language concepts \cite{li2025multimodal}. The Contrastive language-image pretraining (CLIP) framework induces a shared embedding space between images and text. The objective is to pull the embeddings of matched image--text pairs together while pushing apart the embeddings of unmatched pairs drawn from the same training batch. Therefore, a new class can be specified by writing a text prompt instead of curating and labeling a cohort \cite{radford2021clip,luddecke2022image}.

CLIP-based models have been widely adopted in pathology. Pathology VLMs differ in how they initialize their visual and textual representations and learn cross-modal alignment. PathCLIP, QuiltNet, and PathGen-L/14 fine-tune already aligned general-purpose CLIP models on pathology image--text pairs \cite{sun2024pathasst,ikezogwo2023quilt, pathgen2025}. CONCH and MUSK take a different approach, first learning pathology-specific visual and textual representations through unimodal pretraining before aligning them on paired image--text data \cite{lu2024conch,xiang2025vision}.

Nonetheless, these approaches depend on curated image–caption pairs, which are inherently more difficult to obtain than unlabelled slides or biomedical text. Virchow2, for instance, is a self-supervised vision transformer trained on 3.1 million WSIs from diverse institutions, stains, and magnifications, yielding image representations that transfer robustly across pathology tasks \cite{zimmermann2024virchow2,komen2025towards}. On the language side, BioMedBERT is trained from scratch on millions of PubMed abstracts and PubMed Central articles, yielding rich representations of biomedical terminology \cite{gu2021pubmedbert}.

Several models already inherit strong pretrained representations. KEEP initializes its image tower from UNI \cite{chen2024towards} and its text tower from BioMedBERT \cite{gu2021pubmedbert}, then fine-tunes both during vision--language alignment \cite{zhou2026keep}. At the slide level, TITAN trains a slide encoder over frozen CONCH v1.5 tile features, while fine-tuning the slide encoder, text encoder and multimodal decoder during alignment \cite{ding2025multimodal}. PRISM2 aggregates frozen Virchow2 tile embeddings through a 541-million-parameter perceiver trained alongside its language components \cite{vorontsov2026end}. Thus, although these models inherit pretrained representations, cross-modal training still relies on substantial updates to the vision or language components.

Updating large parts of a pretrained model risks overwriting transferable representations acquired during pretraining \cite{lozano2024micro}. Outside pathology, parameter-efficient adaptation has attracted attention partly for this reason \cite{dutt2023parameter,khan2023lilt,lian2025efficient,lai2024bridging}. Within pathology, however, cross-modal alignment has so far relied on substantial updates to the vision or language components. To our knowledge, it has not been examined whether two independently released pretrained vision and text encoders can be aligned into a competitive pathology VLM using only lightweight adapters and projection heads, while leaving both sets of pretrained weights frozen \cite{bafghi2024parameter}.

In this study, we introduce \model, a pathology VLM composed of frozen Virchow2 and BioMedBERT backbones, aligned by attaching rank-4 low-rank adapters (LoRA) and projection heads to both encoders \cite{hu2022lora}. Only the added parameters are trained on the QUILT-1M dataset with a CLIP-style contrastive objective. This means only updating \qty{0.40}{\percent} of the total parameters while the pretrained backbone weights remain frozen throughout. To better understand the trade-off, we also train a matched control that unfreezes both backbones under the same corpus, objective, negative pool and optimizer-step budget.

\model reached the highest mean chance-corrected balanced accuracy across nine public zero-shot patch benchmarks, ahead of PathGen-L/14, KEEP, and CONCH. The matched control that unfroze both backbones showed no detectable advantage in our evaluation. In multi-cancer lymph-node metastasis detection, \model maintained strong performance in nine external cohorts spanning six organs, while other baselines degraded. Cross-modal retrieval was the one axis on which \model did not lead. \model was ranked third by the top-5 retrieval, and second by the margin between matched and random image--caption pairs. 


\section{Results}

\begin{figure}[tb]
    \centering
    \includegraphics[width=1\linewidth]{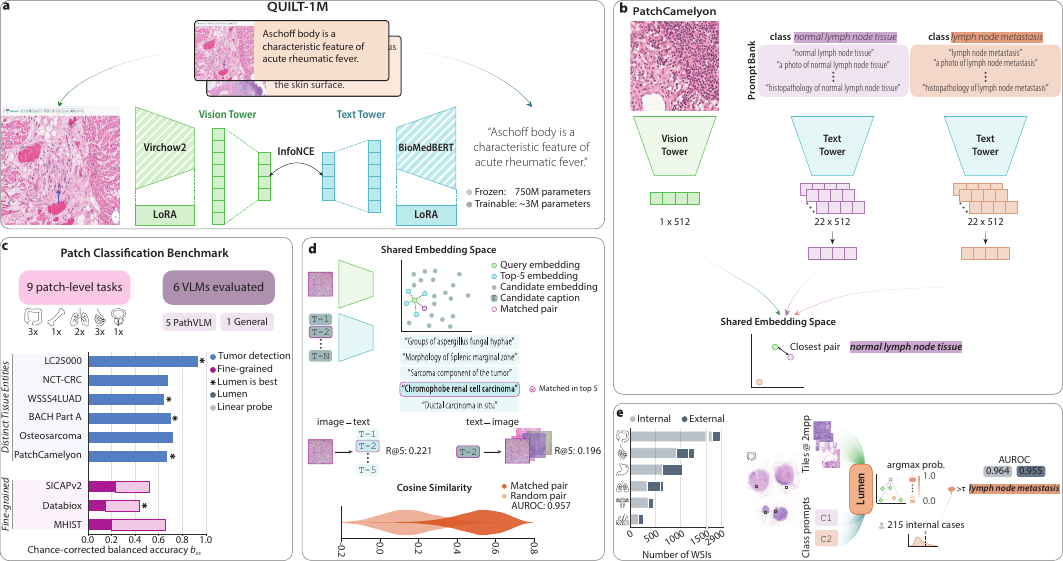}
    \caption{\textbf{Graphical abstract.}
    \textbf{(a)} Model schematic and its training. \model is composed of two independently pretrained frozen encoders, Virchow2 and BioMedBERT, each adapted with rank-4 low-rank adapters and a projection head into a shared 512-dimensional space. The towers are aligned by a symmetric InfoNCE contrastive objective on paired image–caption data from QUILT-1M. Only the adapters and projection heads are trained (0.40\% of the model), while the backbones remain frozen.
    \textbf{(b)} Zero-shot prompt-based classification illustrated on a patch example from PatchCamelyon. A query patch is embedded by the vision tower. Each class name is expanded across a bank of 22 templates, embedded by the text tower, L2-normalized, averaged into a single class vector and L2-normalized again. The predicted label is the class whose vector is nearest the image embedding by cosine similarity in the shared space.
    \textbf{(c)} Patch-classification benchmark. Zero-shot performance of \model across nine public patch-level datasets against six comparison vision–language models. Bars report chance-corrected balanced accuracy ($b_{\mathrm{cc}}$). For each fine-grained dataset, \model's zero-shot score (dark shade) is shown alongside a supervised linear-probe upper bound computed on the same frozen features (light shade). Asterisks mark datasets on which \model is the best-performing VLM.
    \textbf{(d)} Cross-modal retrieval. Top, schematic of retrieval in the shared space: a query image embedding is ranked against candidate caption embeddings, with the matched caption recovered within the top five (highlighted). Middle, bidirectional retrieval, reported as Recall@5 for the image$\rightarrow$text and text$\rightarrow$image directions. Bottom, distribution of cosine similarities for matched versus randomly paired image–caption pairs. Their separation corresponds to a matched-versus-random AUROC of 0.957.
    \textbf{(e)} Slide-level lymph-node metastasis detection.  Left, number of whole-slide images per organ for the internal and external cohorts used for evaluation. Right, whole-slide images are tiled at 2 \textmu m and passed to \model along with the class captions (positive and negative for lymph node metastasis). The cosine similarity between each tile embedding and the two class vectors is computed and converted at the model's learned logit scale to a two-class softmax probability of metastasis. The per-slide score is the maximum tile probability, and a slide is labelled positive when this score exceeds a single threshold $\tau$, fixed by maximizing Youden's index on 215 internal calibration cases and then locked for every cohort (internal AUROC 0.964, external 0.955).}
    \label{fig:graphical_abstract}
\end{figure}

\cref{fig:graphical_abstract} summarizes the approach and scope of evaluation. \cref{fig:graphical_abstract}(a) shows the lightweight alignment of the independently pretrained vision and language towers using LoRA, with QUILT-1M image--text pairs used to optimize the InfoNCE objective. \cref{fig:graphical_abstract}(b) illustrates how the resulting shared embedding space is queried through an ensemble of text prompts for zero-shot classification. These components are described in detail in \cref{sec:training_methods} and \cref{sec:prompt_gen}.

In the main text, we focus on evaluating \model across three complementary axes. \cref{fig:graphical_abstract}(c) summarizes zero-shot patch classification across nine public benchmarks, where \model achieved the highest mean chance-corrected balanced accuracy among the evaluated VLMs. \cref{fig:graphical_abstract}(d) assesses image--text correspondence through cross-modal retrieval and matched-versus-random similarity, while \cref{fig:graphical_abstract}(e) evaluates locked-threshold slide-level lymph-node metastasis detection across internal and external cohorts. We examine each evaluation in detail below.

Alongside these, we report a matched control that unfreezes both backbones under the same corpus, objective, negative pool and optimizer-step budget, evaluated on all three axes under the identical protocol (Supplementary Materials, \cref{sec:fullft}).

\subsection{Public patch benchmarks: robust performance across datasets} \label{sec:patch_bench}

\begin{figure}[!htbp]
    \centering
    \includegraphics[width=0.75\linewidth]{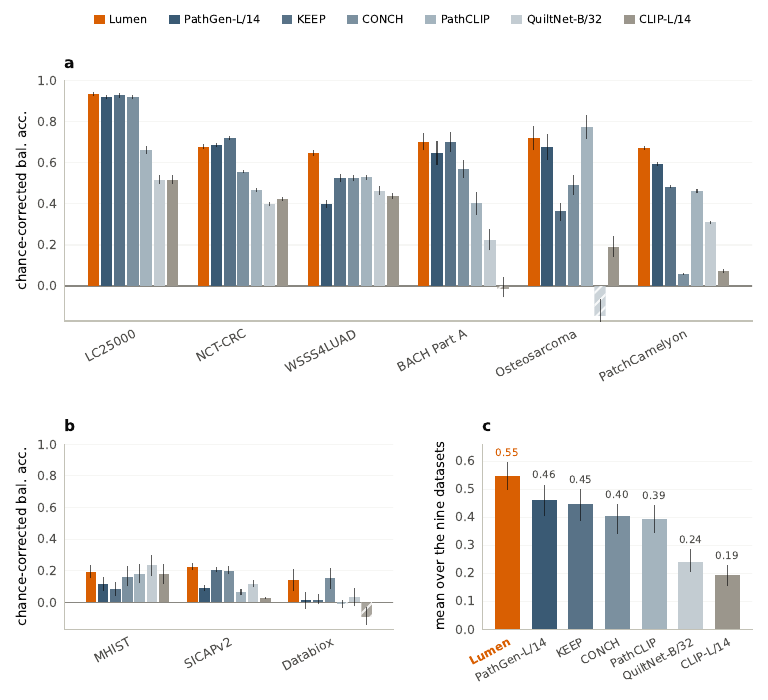}
    \caption{\textbf{Zero-shot patch classification across nine public benchmarks.}
    \textbf{(a)} Chance-corrected balanced accuracy on each of the six datasets whose classes are distinct tissues or entities, with a \qty{95}{\percent} bootstrap interval on every bar. \textbf{(b)} The three fine-grained datasets whose classes are instead adjacent grades or subtypes of one entity, on the same scale as (a). Notably, every model falls close to chance on these. Hatched bars fall below chance, meaning the model selects the wrong class more often than a label-blind predictor. \textbf{(c)} The mean over all nine datasets, one bar per model. The whisker is the range of that mean across the nine leave-one-dataset-out subsets, so it shows how far the ranking depends on any single dataset. All models met the same prompt protocol, one canonical class name across 22 templates, giving 22 prompts per class for every dataset.}
    \label{fig:patch}
\end{figure}

We benchmarked \model against six VLMs on nine commonly used public pathology patch benchmarks under a standardized zero-shot protocol (Methods, \cref{sec:datasets}). All models used their model-native image preprocessing and the same prompt bank, one canonical class name expanded across 22 templates (\cref{sec:prompt_gen}). \cref{fig:patch} reports chance-corrected balanced accuracy for all seven models individually. Macro-AUROC and macro-$F_1$ under the same protocol are given in Supplementary \cref{fig:supp_patch_metrics}.

CLIP-L/14 is the only model in this comparison without pathology-specific or biomedical pretraining. It therefore serves not only as a general-purpose baseline, but also as a measure of how well generic vision--language pretraining transfers to pathology. Several pathology models build directly on OpenAI CLIP checkpoints. QuiltNet-B/32 starts from ViT-B/32, PathCLIP from ViT-B/16, and PathGen-L/14 from CLIP-L/14. Therefore, CLIP-L/14 and PathGen-L/14 form a comparison in our study that isolates the effect of this pathology-specific training.

On the mean chance-corrected scale shown in \cref{fig:patch}(c), \model had the highest observed nine-dataset mean ($0.546$), followed by PathGen-L/14 ($0.461$) and KEEP ($0.448$). Against PathGen-L/14, the strongest baseline on the nine-dataset aggregate, \model's mean paired advantage was $0.086$ (task-bootstrap \qty{95}{\percent} interval, $0.042$--$0.136$), with a higher value on eight of nine datasets. The observed mean paired advantage remained positive when each dataset was omitted in turn, ranging from $0.065$ (WSSS4LUAD omitted) to $0.097$ (NCT-CRC omitted).

CLIP-L/14 achieved the lowest performance, with a mean chance-corrected balanced accuracy of 0.195. It was not uniformly weak, reaching $0.179$ on MHIST, above CONCH, PathGen-L/14 and KEEP. Nevertheless, every pathology-pretrained model family had a higher aggregate performance than CLIP-L/14.

The matched pair quantifies what pathology captions add to the baseline. PathGen-L/14 reaches $0.461$, a gain of $0.266$ over its CLIP-L/14. Comparable gains are observed across the other CLIP-based models when measured against their starting checkpoints: $+0.189$ for QuiltNet-B/32 (ViT-B/32) and $+0.339$ for PathCLIP (ViT-B/16).

The difference between general-purpose and pathology-specific models was smaller than the variation among the pathology models themselves. QuiltNet provides the most informative comparison because it fine-tunes CLIP on QUILT-1M, the same corpus used to align \model. This holds the alignment corpus approximately constant while changing the pretrained encoders and adaptation strategy. Nevertheless, the stronger QuiltNet-B/32 checkpoint achieved a mean chance-corrected balanced accuracy of only $0.240$. This was $0.045$ above CLIP-L/14 but remained substantially below \model ($0.546$).

Fine-grained classification remained difficult for every model. The best corrected score on Databiox was $0.156$, on MHIST $0.235$, and on SICAPv2 $0.226$, showing that separating adjacent histological grades or colorectal polyp subtypes remained challenging. SICAPv2 is the clearest example. \model achieved a corrected score of $0.226$ (\qty{95}{\percent} CI $0.203$--$0.247$), the highest of the seven models, ahead of KEEP ($0.206$) and CONCH ($0.202$). Its four-class macro-$F_1$ was low ($0.319$), yet its quadratic-weighted $\kappa$ reached $0.651$, indicating that it usually identified the correct broad category while confusing neighbouring grades.

This pattern suggests that \model more often confused adjacent Gleason patterns than benign and malignant tissue. Its performance therefore better supports broad cancer detection than precise grade classification. This limitation appears to arise from the read-out rather than the underlying features. A labelled linear probe of \model's own vision tower recovers SICAPv2's Gleason patterns at $0.518$ against $0.226$ from prompting, and averages $0.764$ across the nine datasets against $0.546$ prompted (Supplementary Materials, \cref{sec:probe}).

\subsection{A clinical problem: multi-cancer lymph-node metastasis detection}
\label{sec:ln}

Lymph-node metastasis detection is a clinical task in which N-staging contributes to prognostic stratification and treatment planning. But assessment can require pathologists to review numerous lymph nodes across multiple whole-slide images. In this setting, it is important that performance remains robust under out-of-distribution shifts. We therefore evaluated how well the model transfers to out-of-distribution data when calibrated using only a small internal sample set.

Each model received one decision threshold, calibrated on $215$ patients from the internal dataset and then fixed for all evaluation cohorts. That threshold was applied unchanged to held-out internal patients (4,214 slides from $860$ patients across five organ groups) and to nine independent external cohorts spanning six organs and multiple institutions (2,368 slides across nine external cohorts, Supplementary \cref{tab:ln_cohorts}). No held-out or external label influenced any operating point, and no per-site adjustment was permitted. Tiling, tile scoring, and threshold calibration are detailed in Methods (\cref{sec:ln_methods}).

\begin{figure}[!htbp]
    \centering
    \includegraphics[width=0.75\linewidth]{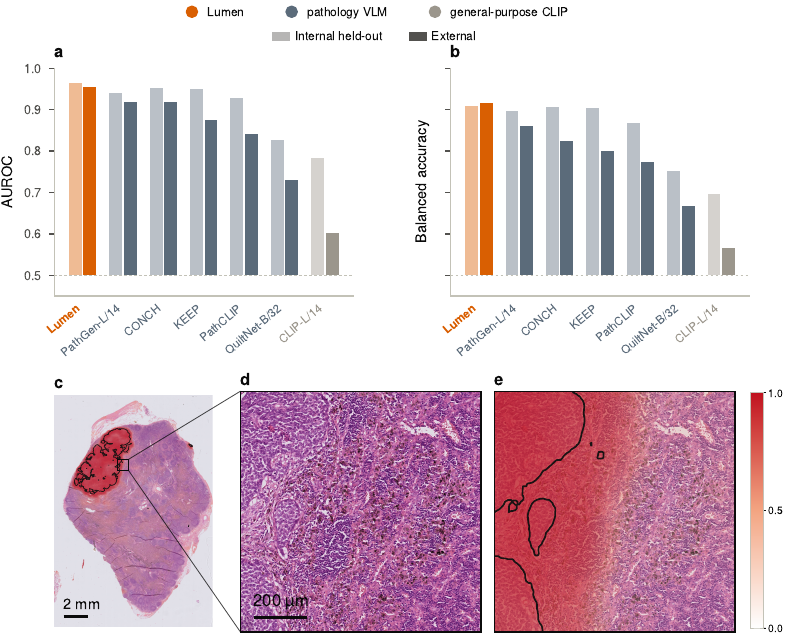}
    \caption{\textbf{Prompt-based lymph-node metastasis detection under one locked decision rule.} \textbf{(a)} Discrimination by AUROC. \textbf{(b)} Balanced accuracy at the locked operating point. In both panels, external performance is in darker shades and internal held-out performance is in lighter shades. Bars rise from the chance level ($0.5$), marked by the dashed line. \textbf{(c)} \model's per-window probability of metastasis on one external slide from the UKK lung cohort, with MetAssist~2.0's tumour outline in black. The box marks the region enlarged in \textbf{(d)} and \textbf{(e)}, shown as H\&E alone in \textbf{(d)} and with prediction overlay in \textbf{(e)}.}
    \label{fig:ln_main}
\end{figure}

\model had the strongest external discrimination of the seven models, with an AUROC of $0.955$ (\qty{95}{\percent} CI $0.942$--$0.966$), ahead of PathGen-L/14 ($0.919$) and CONCH ($0.917$), while the general-purpose baseline was weakest, CLIP-L/14 reaching only $0.603$ (\cref{fig:ln_main}(a)). \model had also the highest external balanced accuracy, $0.915$ (\qty{95}{\percent} CI $0.902$--$0.929$), with sensitivity $0.880$ and specificity $0.951$ (\cref{fig:ln_main}(b)).

From the internal held-out patients to the external cohorts, \model lost only $0.009$ AUROC ($0.964$ to $0.955$) and gained $0.006$ balanced accuracy ($0.909$ to $0.915$), whereas several baselines lost as much as $0.18$ AUROC. This separation is largely a transfer effect and is not visible internally. On the held-out internal patients \model ($0.964$, \qty{95}{\percent} CI $0.956$--$0.971$), CONCH ($0.951$) and KEEP ($0.949$) have overlapping intervals and span only $0.015$. The models separate only on the external cohorts, where \model maintains stronger performance.

A paired bootstrap places \model's pooled external AUROC above every other model's, with a \qty{95}{\percent} interval on the difference that excludes zero even against its two closest competitors, $+0.036$ (\qty{95}{\percent} CI $0.027$--$0.045$) over PathGen-L/14 and $+0.038$ ($0.026$--$0.049$) over CONCH. The unweighted cohort-macro AUROC over the nine external cohorts tells the same story, with \model highest at $0.959$. KEEP is the clearest instance of the transfer gap. Indistinguishable from \model internally, it gives up $0.074$ AUROC on external data, nearly nine times \model's $0.009$.

Across the nine external cohorts, \model's balanced accuracy stayed between $0.88$ and $0.97$ on eight of them. The ninth, lung (HISTAI), fell to $0.70$, because specificity dropped to $0.50$ while sensitivity stayed at $0.90$. Supplementary \cref{fig:ln_supp} reports sensitivity and specificity for every model on every dataset.

We also compared \model with MetAssist~2.0 \cite{garcia2026metassist}, a task-specific end-to-end lymph-node metastasis segmentation model, across \model's held-out cohorts (Supplementary \cref{fig:ln_supp}). The supervised specialist, which operates at very high sensitivity ($0.85$--$1.00$), had higher balanced accuracy on every internal cohort except lung, and on five of the nine external cohorts. \model was higher on the other four, including both melanoma cohorts and the UKK lung cohort, without any lymph-node-specific supervised training. Four of those fourteen cohort-level comparisons are MetAssist~2.0's in-domain results, meaning it was trained on lymph nodes of that organ, and restricting the comparison to the ten where it is generalizing leaves \model higher on four of seven external cohorts. 

\cref{fig:ln_main}(c) shows an example of \model's probability field on one external slide from the UKK lung cohort (Methods, \cref{sec:ln_maps}). The high-probability region aligns with a circumscribed deposit that MetAssist~2.0 also flags independently (depicted by solid outline). \cref{fig:ln_main}(d) shows the corresponding magnified H\&E region, while the same field with the probability overlay in \cref{fig:ln_main}(e) follows the tumour--lymphocyte boundary. The same qualitative comparison on four insightful slides is reproduced in Supplementary Materials (\cref{sec:ln_map_atlas}).

\subsection{Cross-modal retrieval: the limits of alignment}
\label{sec:retrieval}

\begin{figure}[!htbp]
    \centering
    \includegraphics[width=0.75\linewidth]{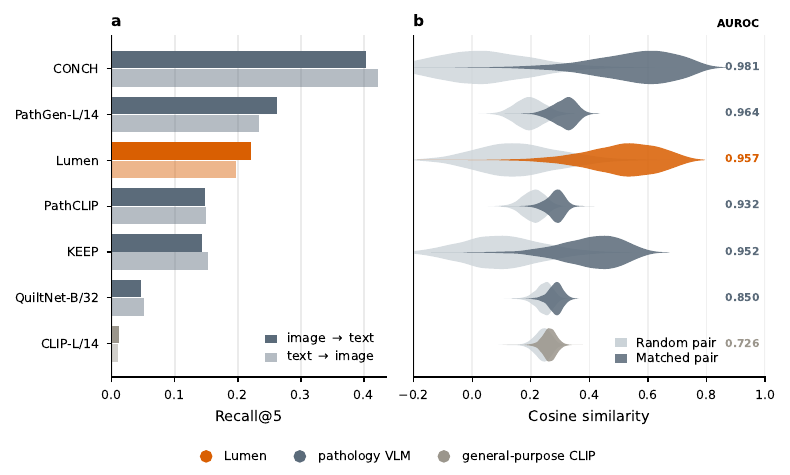}
    \caption{\textbf{Zero-shot cross-modal retrieval on the ARCH dataset.} \textbf{(a)} Recall@5 in both directions, image-to-text (dark shade) and text-to-image (light shade). \textbf{(b)} Cosine-similarity distributions for true image--caption pairs and randomly paired captions. Values at right are the AUROC for separating matched from random pairs.}
    \label{fig:retrieval}
\end{figure}

Classification tests alignment through a small set of predefined class prompts. Retrieval is a harder test of the same correspondence, which we evaluated zero-shot on the ARCH dataset \cite{gamper2021multiple} (Methods, \cref{sec:retrieval_methods}). CONCH achieved the highest image-to-text Recall@5 ($0.403$), ahead of PathGen-L/14 ($0.262$). \model reached $0.221$, above PathCLIP ($0.148$) and KEEP ($0.143$). \cref{fig:retrieval}(a) gives the bidirectional Recall@5 and (b) matched-versus-random AUROC for every model. 

This ordering is not a clean comparison of alignment methods. ARCH is drawn from PubMed figure legends and textbook figures, the native text distribution of CONCH and PathCLIP and a shifted one for models aligned on QUILT-1M or PathGen-1.6M. PathGen-L/14's strong retrieval performance is also consistent with reports that machine-generated captions can benefit cross-modal retrieval \cite{nguyen2023improving}, although its subsequent training on QUILT-1M, PathCap, and OpenPath prevents attributing this effect to synthetic captions alone.

One comparison holds the training data fixed. \model and QuiltNet-B/32 were both aligned on QUILT-1M but used different alignment strategies. \model retrieved the correct caption within the top five nearly five times as often ($0.221$ against $0.046$). \model also outperformed PathCLIP, which was literature-pretrained and evaluated here on the same text distribution ($207$K pairs).

Matched-versus-random separation gives a different view of the same embedding space. \model produced the second-largest mean cosine-similarity margin ($0.351$, behind CONCH's $0.509$) and an AUROC of $0.957$ for distinguishing true from random image--caption pairs (\cref{fig:retrieval}(b)), despite ranking third by image-to-text Recall@5. Therefore, recognizing a random caption--image pair, and ranking the correct caption among thousands of alternative candidates are distinct capabilities.

\section{Discussion} \label{sec:discussion}

Pathology vision--language models do not have to be pretrained end to end. \model combines two independently pretrained encoders, Virchow2 and BioMedBERT, using rank-4 adapters and projection heads while leaving both pretrained backbones frozen. Across nine public patch benchmarks, it ranked first among seven models by mean chance-corrected balanced accuracy. It also transferred a single locked decision rule to external lymph-node slides from six organs without per-site adjustment, outperforming all six VLM comparators on the external transfer task.

The fully unfrozen model optimized the alignment objective more strongly, reaching a lower training loss and higher held-out in-batch retrieval accuracy on QUILT-1M. This advantage did not translate to the downstream benchmarks. On the nine-dataset patch benchmark, the rank-4 model scored $0.546$ compared with $0.507$ for full fine-tuning, although the task-bootstrap \qty{95}{\percent} interval on the difference included zero. The two models were also similar on the held-out internal lymph-node cohort. Their behavior separated on external data. Across the nine external cohorts, the rank-4 model was ahead by $0.013$ AUROC (\qty{95}{\percent} CI $0.005$--$0.020$) and $0.032$ balanced accuracy (\qty{95}{\percent} CI $0.019$--$0.046$). One possible explanation is that updating the backbones weakens invariances inherited from large-scale unimodal pretraining. The retrieval experiment shows the opposite trade-off. Full fine-tuning improved ARCH image-to-text Recall@5 from $0.221$ to $0.251$. Thus, adapting roughly $250$ times more parameters produced a modest retrieval gain but a measurable loss under external distribution shift.

Parameter-efficient alignment also keeps the model modular. The released alignment weights occupy approximately $12$ MB. Such modularity may become increasingly useful as large unimodal pathology models continue to improve and are reused across different multimodal applications.

All three CLIP fine-tuned pathology VLMs evaluated here improved substantially over the checkpoints from which they started. However, even the strongest remained below \model on the patch benchmark. QuiltNet is particularly informative because it was fine-tuned on QUILT-1M, the same corpus used to align \model, yet achieved less than half of \model's mean chance-corrected balanced accuracy. The alignment corpus alone therefore does not explain the difference. The representations brought into alignment also matter.

Our corpus ablation suggests that corpus size alone does not determine alignment quality. PathGen-1.6M contains more image--caption pairs than QUILT-1M, yet it was less effective for aligning the independently pretrained encoders. This does not imply that PathGen-1.6M is a weak pathology corpus: PathGen-L/14 starts from an already aligned CLIP model, is fine-tuned on PathGen-1.6M, and is subsequently trained on an additional \num{700000} pairs from PathCap, QUILT-1M and OpenPath. Rather, caption properties may matter differently when alignment must be learned between two independently pretrained embedding spaces.

The two corpora differ substantially in their textual statistics. Under an identical \num{200000}-token budget using the tokenizer employed during training, PathGen-1.6M contains \num{3153} distinct word pieces compared with \num{8919} for QUILT-1M. Its caption lengths are also much more concentrated, with a median of 63 tokens and a 90th percentile of 70, compared with 88 and 201 tokens for QUILT-1M. These observations are consistent with QUILT-1M providing a more linguistically diverse signal for contrastive alignment. A contrastive objective depends on distinguishing one caption from other captions in the same batch. A corpus with less linguistic variation may therefore provide fewer cues for learning correspondence between two independently pretrained spaces.

A consistent limitation across the experiments appears in the zero-shot read-out. Zero-shot prompting performs well for broad diagnostic categories but is weaker for fine-grained distinctions. A linear probe trained on the same frozen image representations recovers many of the distinctions that prompting misses (Supplementary Materials, \cref{sec:probe}). The visual representation therefore does not appear to be the primary bottleneck on these tasks. Instead, part of the remaining limitation lies in how effectively language can access that representation. This depends jointly on the text encoder, the geometry of the shared embedding space and the prompts used at inference. Improving this language-based read-out is therefore an important direction for future work.

Several aspects remain untested. In the lymph-node metastasis experiment, slide labels record the presence of metastasis but not the size of the deposit. Performance could therefore not be stratified between isolated tumour cells, micrometastases and macrometastases. Extranodal extension, treatment effect and histological subtype were also not evaluated separately.

We showed that effective pathology vision--language modeling does not necessarily require end-to-end retraining of a large multimodal system. \model combines independently pretrained vision and text encoders through lightweight adaptation and achieved strong performance across zero-shot patch classification and multi-cancer lymph-node metastasis detection, although it did not lead on cross-modal retrieval. The matched full-fine-tuning experiment further showed that updating substantially more parameters did not improve downstream performance in our evaluation. Together, these results support parameter-efficient composition as a practical strategy for building pathology VLMs from existing foundation models. As stronger unimodal encoders continue to emerge, whether their improvements translate predictably into stronger composed VLMs remains to be tested.

\section{Methods}\label{sec:methods}

We first describe \model itself (\cref{sec:training_methods}), then the models it is compared against and the prompting procedure common to every evaluation (\cref{sec:models}), then the three evaluations reported in the Results, the patch-level zero-shot benchmark (\cref{sec:patch_methods}), slide-level lymph-node metastasis detection (\cref{sec:ln_methods}) and cross-modal retrieval (\cref{sec:retrieval_methods}), and finally the statistical treatment shared by all of them (\cref{sec:stats}). Analyses reported only in the Supplementary Materials carry their procedure in the supplementary section that reports them.

\subsection{Model architecture and training} \label{sec:training_methods}

\model composes two independently pretrained encoders. Virchow2 \cite{zimmermann2024virchow2} contributes a 2,560-dimensional image embedding, formed by concatenating its class token with the mean of its remaining tokens, and BioMedBERT \cite{gu2021pubmedbert} a 768-dimensional text embedding. Virchow2 places four register tokens between the class token and the patch tokens, and this mean includes them, so it differs slightly from the patch-only pooling given on the backbone's model card. Each embedding is mapped by a single linear layer into a shared 512-dimensional space and layer-normalized, and a learned logit scale completes the contrastive head.

Low-rank adapters \cite{hu2022lora} use rank $r=4$, scaling $\alpha=8$, and dropout $0.1$, with no bias terms, and are inserted into the attention projections of both encoders, that is the fused query--key--value projection and the attention output projection in each of Virchow2's 32 blocks, and the query, key, value, and attention output projections in each of BioMedBERT's 12 layers. This yields 983,040 vision-side and 294,912 text-side adapter parameters together with 1,707,009 parameters in the projection heads and logit scale, giving \num{2984961} trainable parameters in total. All pretrained backbone weights remained frozen throughout, and gradients were propagated only to the adapters and the projection heads.

Training used the standard symmetric Information Noise-Contrastive Estimation (InfoNCE) contrastive objective introduced in CLIP~\cite{radford2021clip} on the QUILT-1M image--text dataset~\cite{ikezogwo2023quilt}, using all \num{653209} stored pairs, of which a fixed \qty{2}{\percent} (\num{13064} pairs, drawn once from a seeded permutation) was held out as the validation set. The contrastive objective is sensitive to the number of in-batch negatives, and it is bounded by activation memory rather than by the trainable parameters. Therefore, we used gradient caching~\cite{gao2021scaling} to increase the number of in-batch negatives. Each optimizer step consumed eight micro-batches of 128 pairs, embedded under \texttt{no\_grad} to assemble the full $\num{1024} \times \num{1024}$ similarity matrix, after which the micro-batches were recomputed with gradients and the cached loss gradients applied. The corpus was stored as a single lightning memory-mapped database (LMDB) of pre-tokenized captions and JPEG bytes.

The model was trained for 20 epochs of 625 optimizer steps, \num{12500} steps in total, with AdamW ($\beta_1 = 0.9$, $\beta_2 = 0.999$, $\epsilon = 10^{-8}$, weight decay $10^{-4}$) under mixed precision, gradients clipped to unit norm. The learning rate warmed up linearly over the first 500 steps to a peak of $5\times10^{-4}$ and then followed a cosine decay to \qty{1}{\percent} of that peak, stepped once per optimizer step. Captions were truncated to 128 tokens and images preprocessed with the same Virchow2 transform used at inference, so the trained checkpoint sees the distribution the benchmark feeds it. The evaluated checkpoint is the one with the lowest held-out loss, reached at epoch 19 of 20 (validation loss $0.9885$, in-batch retrieval accuracy $0.759$ over a 1,024-pair pool).

\subsection{Prompt protocol} \label{sec:models} \label{sec:zeroshot_methods} \label{sec:prompt_gen} \label{sec:zeroshot_scoring}

\model was compared against six released vision--language models, CLIP-L/14 \cite{radford2021clip}, QuiltNet-B/32 \cite{ikezogwo2023quilt}, PathCLIP \cite{sun2024pathasst}, CONCH \cite{lu2024conch}, KEEP \cite{zhou2026keep} and PathGen-L/14 \cite{pathgen2025}. Within a given evaluation every model then met the same class names, the same prompt bank, the same images and the same decision rule. The same protocol governs the matched full fine-tuning control of \cref{sec:fullft}. Our protocol places several baselines below their published values, which were obtained with prompts tuned to each model, so comparisons within this work are internally consistent but are not directly comparable with numbers obtained elsewhere.

Each dataset has one predefined canonical class name per class, fixed in advance and shared unchanged by every model (\cref{tab:canonical_classes}). Class names are expanded across one of two prespecified template banks. The patch benchmark of \cref{fig:patch} uses the 22 prompt frames released with CONCH \cite{lu2024conch}, which include generic image, photomicrograph, histopathology and H\&E formulations, giving 22 prompts per class. The slide-level lymph-node evaluation uses a bank of six templates:

\begin{enumerate}
\item \texttt{\{class\}}
\item \texttt{a photo of \{class\}.}
\item \texttt{an H\&E stained image of \{class\}.}
\item \texttt{histopathology of \{class\}.}
\item \texttt{a histopathology image showing \{class\}.}
\item \texttt{a whole-slide image of \{class\}.}
\end{enumerate}

Each filled prompt was encoded by the model's text encoder and L2-normalized. For each class, prompt embeddings were averaged with equal weight and the mean was L2-normalized again to obtain a single class representation, and image embeddings were compared with the class representations by cosine similarity.

The two evaluations differ in what they do with those similarities. On the patch benchmark the predicted class is the class of highest similarity, an argmax with no free parameter, which is zero-shot in the strict sense. At slide level the tile similarities are instead converted to a two-class softmax probability and read against a model-specific operating threshold calibrated on internal slides (\cref{sec:ln_threshold}), so that evaluation is prompt-based but not zero-shot in the strict threshold-free sense.

\subsection{Zero-shot patch benchmark} \label{sec:patch_methods} \label{sec:datasets}

Patch-level zero-shot classification was evaluated on nine public benchmarks spanning different levels of granularity in tissue typing, tumour detection, grading, and subtype discrimination. \cref{tab:canonical_classes} gives each dataset's source, the size of the evaluated partition, the number of classes and the canonical prompt class names.

Where a dataset defines an official evaluation partition we used it, namely the Kaggle test directory for LC25000, the CRC-VAL-HE-7K cohort for NCT-CRC, the test partition of the staged distribution for Osteosarcoma, and the official test partition for MHIST. Two datasets required a further choice. WSSS4LUAD was restricted to the 4,693 pure single-label patches of the labelled training partition. Databiox releases each specimen at four magnifications, and we evaluated the $40\times$ subset only, without pooling magnifications.

Two datasets contain far fewer independent specimens than images. The 283 Osteosarcoma tiles come from three cases (145, 132 and 6 tiles), and the 454 Databiox images from 124 distinct patient or specimen identifiers, a mean of $3.7$ images each. Neither partition is defined at case level, so image-level intervals on both reflect fewer independent specimens than images (\cref{sec:stats}).

\begin{table}[t]
\centering
\footnotesize
\setlength{\tabcolsep}{3.5pt}
\caption{Patch benchmark datasets, with the size of the evaluated partition and the canonical class names used to build every model's prompts.}
\label{tab:canonical_classes}
\begin{tabular}{lrll}
\toprule
Dataset & Images/Patches & \#classes & Canonical prompt class names \\
\midrule

LC25000 \cite{borkowski2019lc25000} & \num{2499} & 5 &
\begin{tabular}[t]{@{}l@{}}
benign lung tissue \\
lung adenocarcinoma \\
lung squamous cell carcinoma \\
benign colon tissue \\
colon adenocarcinoma
\end{tabular}
\\
\addlinespace

\begin{tabular}[t]{@{}l@{}}NCT-CRC /\\CRC-VAL-HE-7K \cite{kather2019predicting}\end{tabular} & \num{7180} & 9 &
\begin{tabular}[t]{@{}l@{}}
adipose tissue \\
background \\
debris \\
lymphocytes \\
mucus \\
smooth muscle \\
normal colon mucosa \\
colorectal cancer-associated stroma \\
colorectal adenocarcinoma epithelium
\end{tabular}
\\
\addlinespace

WSSS4LUAD \cite{wsss4luad} & \num{4693} & 3 &
\begin{tabular}[t]{@{}l@{}}
lung adenocarcinoma tumor epithelium \\
lung adenocarcinoma tumor-associated stroma \\
normal lung tissue
\end{tabular}
\\
\addlinespace

Osteosarcoma \cite{foley2020osteosarcoma} & $283$ & 3 &
\begin{tabular}[t]{@{}l@{}}
non-tumor tissue \\
necrotic tumor \\
viable tumor
\end{tabular}
\\
\addlinespace

PatchCamelyon \cite{veeling2018rotation} & \num{32768} & 2 &
\begin{tabular}[t]{@{}l@{}}
normal lymph node tissue \\
lymph node metastasis
\end{tabular}
\\
\addlinespace

BACH Part A \cite{aresta2019bach} & $400$ & 4 &
\begin{tabular}[t]{@{}l@{}}
normal breast tissue \\
benign breast lesion \\
breast carcinoma in situ \\
invasive breast carcinoma
\end{tabular}
\\
\addlinespace

SICAPv2 \cite{silva2020sicapv2} & \num{2122} & 4 &
\begin{tabular}[t]{@{}l@{}}
benign prostate tissue \\
Gleason grade 3 prostate cancer \\
Gleason grade 4 prostate cancer \\
Gleason grade 5 prostate cancer
\end{tabular}
\\
\addlinespace

Databiox \cite{bolhasani2020histopathological} & $454$ & 3 &
\begin{tabular}[t]{@{}l@{}}
grade 1 invasive ductal carcinoma \\
grade 2 invasive ductal carcinoma \\
grade 3 invasive ductal carcinoma
\end{tabular}
\\
\addlinespace

MHIST \cite{wei2021mhist} & $977$ & 2 &
\begin{tabular}[t]{@{}l@{}}
hyperplastic colorectal polyp \\
sessile serrated adenoma
\end{tabular}
\\
\bottomrule
\end{tabular}
\end{table}

\subsection{Slide-level lymph-node metastasis detection} \label{sec:ln_methods}

\subsubsection{Cohorts and labels}

Slide-level lymph-node metastasis detection was evaluated on internal and external whole-slide image cohorts. The internal cohort comprised lymph-node slides from the Institute of Tissue Medicine and Pathology, University of Bern, spanning breast, colorectal, endometrial, lung, and upper-gastrointestinal cancers. External cohorts included CAMELYON17 \cite{litjens20181399}, HISTAI cohorts across multiple tumour types \cite{histai_data}, a lung cancer cohort from the University Hospital of Cologne, and a public gastric lymph-node dataset \cite{uppergi_data}. Cohort sizes and positive/negative splits are reported in Supplementary \cref{tab:ln_cohorts}.

For the internal cohorts, slide labels were derived from dedicated slide review performed by expert pathologists following AJCC criteria. Slides were digitised on Pannoramic 1000 DX (3D 3DHISTECH, Budapest, Hungary) scanners at $40\times$ magnification.

The internal cohort was split by patient into a \qty{20}{\percent} calibration set and an \qty{80}{\percent} held-out test set, stratified by organ and by whether a patient's slides were negative or positive. The same calibration split was shared by all evaluated models, and the external cohorts were held out in full.

\subsubsection{Tiling and slide scoring} \label{ln_scoring}

Slides were tiled at 2.0 microns per pixel using non-overlapping $224 \times 224$ tiles. Tiles were retained with an HSV-based tissue filter, and each retained tile was encoded to a normalized image embedding.

Three predefined negative--positive class-name pairs were each expanded across the six canonical templates of \cref{sec:prompt_gen}, giving an 18-prompt ensemble per polarity, combined by the procedure of \cref{sec:zeroshot_scoring}. The three pairs were:

\begin{enumerate}
    \item \texttt{normal lymph node tissue}; \texttt{lymph node tissue with metastatic carcinoma}
    \item \texttt{a lymph node negative for metastatic carcinoma}; \texttt{a lymph node positive for metastatic carcinoma}
    \item \texttt{healthy lymph node tissue}; \texttt{metastatic tumor in lymph node tissue}
\end{enumerate}

Tile-level cosine similarities to the two resulting class vectors were converted to a two-class softmax probability, and the continuous slide score was the maximum positive-class probability across retained tiles. Two aspects of this read-out were varied as sensitivity checks, the carcinoma wording of the shared positive vocabulary on the two melanoma cohorts and the use of the maximum as the slide aggregator. Neither changes the result, and both are reported in Supplementary \cref{sec:readout_checks}.

\subsubsection{Threshold calibration and metrics} \label{sec:ln_threshold}

For each model, a single operating threshold was selected on the internal calibration slides by maximizing Youden's index, $J = \mathrm{Sensitivity} + \mathrm{Specificity} - 1$. Balanced accuracy is the mean of sensitivity and specificity, so it is a monotone function of Youden's index and the two are maximized by the same threshold. The highest threshold attaining the maximum Youden index was retained. This model-specific threshold was then locked and applied unchanged to the internal held-out patients and all external cohorts. AUROC was computed from the continuous slide scores, and balanced accuracy, sensitivity, specificity, and $F_1$ were computed at the locked threshold.

\subsubsection{Slide-level probability maps} \label{sec:ln_maps}

The maps in \cref{fig:ln_main}(c,e) is a visualization and follow a different protocol from the score. The slide is re-read at $2.0$ micron per pixel and swept with a $224$ px window at stride $28$, giving \qty{88}{\percent} overlap. Each window's positive-class probability is computed at the model's learned logit scale, then soft-voted onto a grid of $4$ microns per pixel with a Gaussian kernel ($\sigma = 0.125$ of the window). Pixels reached by no window are masked with the background.

\subsection{Cross-modal retrieval} \label{sec:retrieval_methods}

Cross-modal retrieval was evaluated on ARCH \cite{gamper2021multiple}, a held-out set of \num{7576} pathology figure--caption pairs from PubMed articles and textbooks. Each model embedded the images and captions with its frozen encoders, ranked them by cosine similarity, and we report bidirectional Recall@5. Because \num{1083} of the \num{7576} captions are shared by more than one figure, Recall@5 was also recomputed treating identical captions as a single candidate, which changes neither the values nor the ordering appreciably (Supplementary \cref{sec:readout_checks}).

Coarse image--text correspondence was quantified separately by randomly permuting the captions once with random-number seed 0, excluding identity pairings, and computing the AUROC that separates the \num{7576} true-pair cosine similarities from the \num{7576} randomly paired similarities. We also report the cosine-similarity margin, defined as the mean true-pair similarity minus the mean random-pair similarity. This analysis asks whether a matching pair scores above an unrelated pair. Unlike Recall@5, it does not require the true caption to outrank every near-miss in the retrieval set.

\subsection{Statistics and reproducibility} \label{sec:stats} \label{sec:patch_metrics}

\paragraph{Metrics.} The primary patch-level metric was chance-corrected balanced accuracy,
\[
b_{\mathrm{cc}} = \frac{b-c}{1-c},
\]
where $b$ is the raw balanced accuracy, $K$ is the number of classes, and $c=1/K$ is the expected balanced accuracy of a label-blind predictor. This transformation places chance at $0$ and perfect classification at $1$, so datasets with different numbers of classes are placed on a common scale. Within a dataset it is a monotone affine transformation of balanced accuracy and therefore does not change the ordering of models on that dataset. Each model's patch-benchmark score is its mean chance-corrected balanced accuracy across the nine datasets. Macro-AUROC and macro-$F_1$ are reported as secondary metrics, macro-$F_1$ as the unweighted mean of the per-class $F_1$ scores. For binary tasks, AUROC was calculated from the positive-class softmax probability. For multiclass tasks, macro-AUROC was computed in a one-versus-rest manner when all required classes were present.

\paragraph{Resampling unit.} All intervals reported in this work are bootstrap percentile intervals at the 2.5th and 97.5th percentiles. The resampling unit is the largest unit of statistical independence the data supports. For the lymph-node evaluation that is the patient or case, with one exception, the public gastric cohort, which carries no patient or case labels, so there we resampled individual slides. Most of the patch datasets carry no patient identifiers, so there the unit is the image, and intervals understate uncertainty where images share a specimen. Osteosarcoma and Databiox are the two clearest instances (\cref{sec:datasets}).

\paragraph{Model comparisons.} Paired comparisons on the patch benchmark resample the nine datasets rather than the patches, with both models scored on the same resample, \num{10000} resamples at seed 0. Each comparison was repeated nine times, omitting one dataset in turn, and \cref{fig:patch}(c) reports the resulting range. The per-dataset intervals in \cref{fig:patch}(a,b) are within-dataset precision rather than generalization, from \num{1000} resamples of that dataset's images. Paired comparisons on the lymph-node benchmark use $978$ external resampling units: patients or cases where identifiers are available, and individual slides for the public gastric cohort, which provides no patient or case identifiers. Where values within a dataset are not independent, as for the cross-validation folds of the supplementary linear probe, the nine per-dataset mean differences were tested with a Wilcoxon signed-rank test \cite{Wilcoxon1945} rather than the individual folds.


\section*{Acknowledgements}

The authors would like to thank the Translational Research Unit (TRU) of the ITMP for their assistance staining and scanning slides and the UBELIX team (the HPC cluster at the University of Bern), for providing and maintaining the computational resources needed to conduct this study.

\section*{Ethics Approval and Consent to Participate}

Responsible ethics committees granted individual approval for each collected pseudoanonymized cohort. All cohorts from Bern were approved by the Ethics Committee of the Canton of Bern (reference number: b2021-00033) in accordance with the Human Research Act HFG 2014. The cohort from the University Clinic of Cologne was generated within the FED-PATH project and approved by the University of Cologne ethics board (22-1233 and 20-1583).

\section*{Funding Statement}
This study was supported by the Swiss National Science Foundation (10.000.619), the Center for Artificial Intelligence in Medicine (CAIM), now the Department of Digital Medicine, University of Bern and ISREC Cancer Research Foundation, Lausanne.

\section*{Author Contributions}
Author contributions are listed below in the order of appearance in the author list.

K.T. trained and evaluated the model, performed the ablation experiments, produced the figures, and wrote the manuscript.
M.J. and A.F. contributed to model training, evaluation, and ablation experiments.
J.G.-B. contributed case selection, and digital cohort curation for the Bern lymph-node datasets, contributed to the figures, and writing the manuscript.
P.Z., B.Z., and B.D. contributed case selection, slide evaluation, and pathological annotation for the Bern lymph-node cohorts.
M.D.B. and A.P. provided clinical and scientific support and reviewed the manuscript.
Y.T. provided the lung cohort with expert slide-level label annotations from the University Clinic of Cologne and reviewed the manuscript.
I.Z. provided resources, and reviewed the manuscript.
A.K. conceived, designed, and supervised the project and contributed to writing the manuscript.

\section*{Data availability}
The public benchmark datasets used in this study are available from their original repositories. Internal lymph node datasets are not publicly available due to privacy and ethics restrictions, but may be made available upon reasonable request, subject to appropriate ethical and institutional approvals.

\section*{Code availability}
The trained low-rank adapters and projection heads that constitute \model's alignment (\num{2984961} parameters, \qty{0.40}{\percent} of the model) are released as open weights on the Hugging Face Hub at \url{https://huggingface.co/digitalpathologybern/Lumen}, and the frozen Virchow2 and BioMedBERT backbones they attach to are obtained from their original public releases.

\section*{Disclosure}
P.Z. has received honoraria from AstraZeneca.

\bibliography{sn-bibliography}%

\pagebreak


\clearpage

\begin{center}
    {\Large \papertitle\par}

    \vspace{1.2em}

    {\large\bfseries Supplementary Materials\par}
\end{center}

\vspace{2em}

\setcounter{section}{0}
\setcounter{subsection}{0}
\setcounter{subsubsection}{0}

\renewcommand{\thesection}{S\arabic{section}}
\renewcommand{\thesubsection}{\thesection.\arabic{subsection}}
\renewcommand{\thesubsubsection}{\thesubsection.\arabic{subsubsection}}

\setcounter{figure}{0}
\setcounter{table}{0}
\renewcommand{\thefigure}{S\arabic{figure}}
\renewcommand{\thetable}{S\arabic{table}}

\section{Parameter-efficient vision-language alignment ablation}
\label{sec:lora_alignment}

To isolate which parameters must be adapted for cross-modal alignment, we compared four configurations that share the same encoders, projection heads, contrastive objective, and training data. They differ only in where low-rank adaptation is applied: (i)~projection heads only, with both encoders frozen; (ii)~LoRA on the text encoder only; (iii)~LoRA on the vision encoder only; and (iv)~LoRA on both encoders. Learning is restricted to low-rank updates inside the encoders as well as the projection heads.

The ablation models are trained with an InfoNCE objective on paired histopathology image--caption data from QUILT-1M \cite{ikezogwo2023quilt}. All four configurations were evaluated zero-shot on three held-out patch tasks: the LC25000 colon (two-class) and lung (three-class) subsets and PatchCamelyon lymph-node metastasis (\cref{tab:lora_placement}). These configurations were scored with a single prompt per class rather than the prompt ensembles used in the main evaluations. Furthermore, the LC25000 lung and colon subsets are scored separately here rather than as the single five-class task of the main benchmark. Values in this table are therefore internally comparable across the four rows but not comparable with the main patch benchmark.

Two of the datasets used to make this choice, LC25000 and PatchCamelyon, also appear in the nine-dataset patch benchmark reported in the main text. The aggregate over those nine datasets is, therefore, not fully independent of the configuration chosen here. Two things limit how far that reaches. The choice was between four configurations that differ only in which parts of the network are adapted, none of them tuned to a dataset, and the protocol differs from the benchmark as set out above. Most directly, the result does not depend on those two datasets. Removing both and recomputing over the remaining seven leaves \model first with a mean chance-corrected balanced accuracy of $0.473$ against $0.381$ for the strongest remaining baseline, and a paired advantage over PathGen-L/14 of $0.097$ (95\% CI $0.044$ to $0.158$) against $0.086$ ($0.042$ to $0.136$) over all nine. The margin is slightly larger without them, not smaller.

\begin{table*}[h]
    \centering
    \caption{Zero-shot top-1 accuracy (\%) for four adaptation configurations. ``Full model'' is the trainable share of about 741 million backbone parameters plus each configuration's adapters and heads.}
    \label{tab:lora_placement}
    \renewcommand{\arraystretch}{1.15}
    \scriptsize
    \setlength{\tabcolsep}{3pt}
    \begin{tabular}{lrrrrr}
        \toprule
        \textbf{Adapted parameters} & \textbf{Trainable} & \textbf{Full model} & \textbf{LC-Colon} & \textbf{LC-Lung} & \textbf{PCam} \\
        \midrule
        Projection                                     & 1,707,009 & \qty{0.23}{\percent} & 95.0 & 76.0 & 60.0 \\
        LoRA: text encoder                             & 2,001,921 & \qty{0.27}{\percent} & 100.0 & 87.7 & 74.0 \\
        LoRA: vision encoder                           & 2,690,049 & \qty{0.36}{\percent} & 99.0 & 94.2 & 76.7 \\
        \textbf{LoRA: both encoders}         & \textbf{2,984,961} & \textbf{\qty{0.40}{\percent}} & \textbf{100.0} & \textbf{96.0} & \textbf{84.0} \\
        \bottomrule
    \end{tabular}
\end{table*}

Trainable-parameter counts include both direct projection heads, the learned contrastive logit scale, and the LoRA parameters enabled in each configuration. With rank $r=4$, the vision-side adapters contribute 983,040 parameters and the text-side adapters contribute 294,912 parameters. The projection heads and logit scale contribute 1,707,009 parameters in every configuration.

As shown in \cref{tab:lora_placement}, performance improved as more of the network was adapted. On the near-saturated colon task, training only the projection heads achieved \qty{95.0}{\percent} accuracy, while all LoRA configurations reached essentially \qty{100}{\percent}. The effect was more pronounced on the harder three-class lung task and especially on PatchCamelyon. On PatchCamelyon, adapting either the text encoder (\qty{74.0}{\percent}) or the vision encoder (\qty{76.7}{\percent}) outperformed training only the projection heads (\qty{60.0}{\percent}), while adapting both encoders achieved the best performance (\qty{84.0}{\percent}). Across all three datasets, vision-only adaptation consistently outperformed text-only adaptation, although neither matched adapting both encoders.

\section{Full fine-tuning control}
\label{sec:fullft}

The alignment above trains \num{2984961} parameters, \qty{0.40}{\percent} of the composed model. To ask what that restriction costs, we trained a second model that differs from it in one respect. Both backbones were unfrozen, giving \num{742428673} trainable parameters. The corpus, the InfoNCE objective, the \num{1024}-pair negative pool, the \num{12500} optimizer steps, the head learning rate and the gradient-clipping norm were all held fixed at the values of the rank-4 run (\cref{sec:training_methods}).

Four settings necessarily differ, because a rate that trains a randomly initialised projection head destroys a pretrained transformer. Backbone weights take a learning rate of $1\times10^{-5}$ against $5\times10^{-4}$ for the heads, warm-up is extended to \num{1000} steps, $\beta_2$ is reduced to $0.95$, and no weight decay is applied to normalisation gains, biases or the logit scale. Mixed precision uses bfloat16 rather than float16. The control was evaluated on the patch benchmark, the linear probe, cross-modal retrieval and the slide-level lymph-node benchmark.

Full fine-tuning fits the training objective far harder, reaching a lower training loss and a higher held-out in-batch retrieval accuracy on QUILT-1M than the rank-4 model attains at any point. It does not transfer better. On the nine-dataset patch benchmark under the same prompt protocol used in \cref{fig:patch}, the rank-4 model scored $0.546$ against $0.507$ for full fine-tuning on the chance-corrected scale, a paired advantage of $0.040$ whose task-bootstrap \qty{95}{\percent} interval ($-0.011$ to $0.102$) includes zero. The two are therefore not separable, and adapting $250$ times more parameters did not produce a detectable gain in our setting.

The lymph-node benchmark separates them where the patch benchmark does not. On the held-out internal patients the two are within $0.011$ on both AUROC and balanced accuracy, but on the nine external cohorts the rank-4 model is ahead by $0.013$ AUROC (\qty{95}{\percent} CI $0.005$--$0.020$) and $0.032$ balanced accuracy ($0.019$--$0.046$), from a paired bootstrap over the 978 external resampling units. Full fine-tuning is therefore not worse in domain and is worse out of it, which is what would be expected if updating the backbones erodes an invariance the pretrained encoders bring and the alignment corpus is too narrow to restore.

Two secondary measurements favour full fine-tuning slightly, which is worth stating because it bounds what the low-rank constraint costs. When linearly probed against stock Virchow2 on identical folds, the fully fine-tuned vision tower is the better representation ($+0.016$ against $+0.006$ chance-corrected), and it sits closer to the stock representation than the rank-4 tower does, by both cosine similarity and relative $L_2$ displacement. Adapting every weight gently perturbs the representation less than a rank-4 update scaled by $\alpha/r = 2$ at every attention projection. Cross-modal retrieval on ARCH is also better (Recall@5 $0.251$ against $0.221$). Neither advantage reaches the prompted benchmark. Across the nine datasets the change in linear probe accuracy and the change in zero-shot accuracy are uncorrelated (Pearson $r = -0.29$, $p = 0.45$).

\section{Training corpus ablation} \label{sec:corpus_ablation}

The ablation in \cref{sec:lora_alignment} varies which parameters are adapted while holding the corpus fixed. Here we hold everything else fixed and vary the corpus, replacing QUILT-1M with PathGen-1.6M \cite{pathgen2025}. PathGen-1.6M is the larger of the two: 1,606,064 image--caption pairs cut from 7,203 TCGA whole-slide images, each caption written by a captioning model from the patch it describes, against QUILT-1M's 653,209 pairs transcribed from educational videos.

Both arms use the same encoders, the same rank-4 adapters on both encoders, the same projection heads, the same contrastive objective with a 1,024-pair negative pool, and the same optimizer and schedule. They are also matched on optimizer steps: 12,500 for QUILT-1M and 13,125 for PathGen-1.6M, so each model saw approximately $1.3 \times 10^{7}$ pairs. Because PathGen-1.6M is $2.5\times$ larger, this is 20 passes over QUILT-1M against 8.4 over PathGen-1.6M. Validation for the PathGen-1.6M arm holds out whole slides rather than random patches.

Under an identical evaluation protocol the PathGen-1.6M-trained model is worse on every benchmark in this work (\cref{tab:corpus_ablation}), and by large margins: $0.181$ balanced accuracy on the patch benchmark and $0.147$ AUROC on lymph-node metastasis detection, with cross-modal retrieval falling by a factor of five.

\begin{table}[t]
\centering
\small
\caption{Comparison of the QUILT-1M and PathGen-1.6M training corpora across the experiments reported in this manuscript. Values are reported in their native metric (balanced accuracy, AUROC, Recall@1, or median rank). The linear probe uses the projected 512-dimensional image embeddings. It is therefore comparable only between the two models shown here. Better value is indicated in bold.}
\label{tab:corpus_ablation}
\renewcommand{\arraystretch}{1.15}
\setlength{\tabcolsep}{6pt}
\begin{tabular}{lccr}
    \toprule
    \textbf{Evaluation} & \textbf{QUILT-1M} & \textbf{PathGen-1.6M} & \textbf{$\Delta$} \\
    \midrule
    \multicolumn{4}{l}{\textit{Zero-shot, uses image and text towers}} \\
    Patch benchmark (mean balanced acc.)   & \textbf{0.683} & 0.502 & $-0.181$ \\
    Lymph node (held-out AUROC)            & \textbf{0.964} & 0.817 & $-0.147$ \\
    ARCH retrieval (i2t R@1)               & \textbf{0.089} & 0.016 & $-0.073$ \\
    ARCH retrieval (t2i R@1)               & \textbf{0.071} & 0.014 & $-0.057$ \\
    ARCH retrieval (t2i median rank)       & \textbf{41}    & 447   & $+406$ \\
    \midrule
    \multicolumn{4}{l}{\textit{Linear probe, uses the image tower only}} \\
    Mean over the nine patch datasets      & 0.895 & 0.888 & $-0.007$ \\
    \bottomrule
\end{tabular}
\end{table}

The linear probe helps locate the source of the failure. The linear probe reaches $0.895$ for the QUILT-1M arm and $0.888$ for the PathGen-1.6M arm. PathGen-1.6M is even stronger on osteosarcoma ($0.952$ vs. $0.933$) and WSSS4LUAD ($0.974$ vs. $0.964$). Its image representations are therefore essentially as strong as those learned with QUILT-1M. The difference appears in the image-text alignment. The gap between linear-probe and zero-shot accuracy is $0.212$ for QUILT-1M, but $0.386$ for PathGen-1.6M.

This does not imply that PathGen-1.6M is a weak corpus. The released PathGen-L/14 models instead suggest that it provides useful pathology-specific visual grounding, but not enough diagnostic language to make that representation fully accessible through text. PathGen-L/14 starts from an OpenAI CLIP model, is fine-tuned on PathGen-1.6M, and is then trained on another \num{700000} pairs from PathCap, QUILT-1M and OpenPath. The authors motivate this second stage by noting that the generated PathGen-1.6M captions describe morphology but often lack the diagnostic language needed to anchor those descriptions \cite{pathgen2025}. The additional corpora, including QUILT-1M, provide that missing language.

Our corpus ablation reaches the same conclusion from a different direction. The linear probe shows that PathGen-1.6M learns a strong image representation, but the much larger zero-shot gap shows that the correspondence to language is weaker. This matters especially in our setting, because Virchow2 and BioMedBERT are pretrained independently on images and text. There is no existing multimodal alignment to fall back on, so that correspondence must be learned entirely from the training captions. In this setting, QUILT-1M is the better training corpus.

\section{Linear probe of the vision tower}
\label{sec:probe}

Every zero-shot result in the main text reads the vision tower through text, which leaves two questions open: when a prompted model fails, is the visual information absent or only unreachable through language, and does aligning the tower to text cost it performance? We read the same tower a second way, with labels instead of prompts, fitting a multinomial logistic regression to its \num{2560}-dimensional output under cross-validation. This is deliberately not a fair contest. The linear probe sees labels and the prompt does not, so it bounds what is present rather than what is reachable.

Features were taken at the output of the vision tower, that is the \num{2560}-dimensional concatenation of the class token with the mean of the remaining tokens (Methods, \cref{sec:training_methods}), before the projection head. The head is trained, so probing after it would confound the quality of the visual features with the quality of the head. Features were L2-normalized and classified with multinomial logistic regression ($C = 1$, at most \num{2000} iterations) under five-fold cross-validation, reduced to as many folds as a dataset has groups when it has fewer than five (osteosarcoma, with three cases, is therefore scored with three folds), scored by balanced accuracy and reported on the same chance-corrected scale as the prompt-based results.

Folds were grouped by biological source wherever the source is recoverable from the sample identifier, so correlated patches from one case, slide, WSI or specimen cannot fall on both sides of a split. Osteosarcoma was grouped by case (3 groups), SICAPv2 and WSSS4LUAD by slide/WSI (31 and 50), and Databiox by specimen (124). For the remaining datasets (LC25000, PatchCamelyon, MHIST, BACH Part A, NCT-CRC) no shared source is recoverable from the identifiers, so ordinary stratified folds were used. Because folds from the same dataset are not independent replicates, statistical testing was performed on the nine per-dataset mean differences rather than on individual folds. We tested these paired differences using a Wilcoxon signed-rank test. The adapted tower is compared with the stock Virchow2 tower on identical folds. Feature displacement between the towers is reported as the per-sample cosine similarity and the norm-relative change $\lVert f_\text{on} - f_\text{off}\rVert / \lVert f_\text{off}\rVert$.

According to \cref{tab:vision_probe} it is present, and the gap is large. Averaged over the nine datasets, prompting \model reaches $0.546$ chance-corrected balanced accuracy while a linear probe of its own tower, cross-validated with folds grouped by biological source wherever the source is recoverable, reaches $0.764$.

The gap is not spread evenly. On the three tasks this work describes as remaining difficult, the linear probe recovers far more than prompting: SICAPv2's Gleason patterns at $0.518$ against $0.226$ from prompting, MHIST's polyp subtypes at $0.656$ against $0.194$, and Databiox's grades at $0.433$ against $0.141$, where no evaluated model exceeds $0.156$ by prompting.

One dataset runs the other way. Osteosarcoma has only three cases, and under case-grouped folds the linear probe generalizes worse than prompting ($0.478$ against $0.722$), which says that with so few specimens neither read-out is reliable there, not that the features lack the distinction.

Aligning the tower to text showed no evidence of degrading its image-only performance. The adapted tower performed slightly better on average ($+0.006$ chance-corrected balanced accuracy) and outperformed the original tower on seven of nine datasets, although the difference was not statistically significant (Wilcoxon signed-rank $p = 0.30$). The adapters are nonetheless doing substantial work, with adapted and stock features rotated by about $55$ degrees with a \qty{15}{\percent} reduction in norm, so the alignment is not a marginal perturbation of the encoder. 

\begin{table}[t]
\centering
\small
\caption{A labelled read-out of \model's own vision tower, against its prompted read-out. All values are chance-corrected balanced accuracy. \emph{Headroom} is their difference, what the prompt-based read-out does not recover from a representation that demonstrably contains it. \emph{$\Delta$ tower} is the paired per-fold difference between \model's LoRA-adapted tower and the stock Virchow2 backbone recovered from the same checkpoint, positive meaning the adapted tower is better.}
\label{tab:vision_probe}
\renewcommand{\arraystretch}{1.15}
\setlength{\tabcolsep}{6pt}
\begin{tabular}{lcccc}
    \toprule
    \textbf{Dataset} & \textbf{Zero-shot} & \textbf{Linear probe} & \textbf{Headroom} & \textbf{$\Delta$ tower} \\
    \midrule
    NCT-CRC          & $0.678$ & $0.999$ & $+0.321$ & $+0.002$ \\
    LC25000          & $0.934$ & $0.995$ & $+0.061$ & $+0.002$ \\
    PatchCamelyon    & $0.673$ & $0.945$ & $+0.272$ & $+0.008$ \\
    BACH Part A      & $0.703$ & $0.940$ & $+0.237$ & $+0.007$ \\
    WSSS4LUAD        & $0.647$ & $0.913$ & $+0.266$ & $-0.009$ \\
    Osteosarcoma     & $0.722$ & $0.478$ & $-0.244$ & $-0.065$ \\
    \midrule
    MHIST            & $0.194$ & $0.656$ & $\mathbf{+0.462}$ & $+0.029$ \\
    SICAPv2          & $0.226$ & $0.518$ & $\mathbf{+0.292}$ & $+0.014$ \\
    Databiox         & $0.141$ & $0.433$ & $\mathbf{+0.292}$ & $+0.069$ \\
    \midrule
    \textbf{Mean}    & $\mathbf{0.546}$ & $\mathbf{0.764}$ & $\mathbf{+0.218}$ & $+0.006$ \\
    \bottomrule
\end{tabular}
\end{table}

\section{Prompt-protocol robustness} \label{sec:prompt_protocol}

A zero-shot score is a property of a model and a prompt protocol jointly, so any ranking has to be checked against a protocol the ranking did not choose. We re-scored the entire patch benchmark, all seven models on all nine datasets, under the six-template bank of \cref{sec:prompt_gen} in place of the 22-template bank reported in \cref{fig:patch}. Both banks use the same canonical class names, so only the template set differs.

Phrasing turns out not to matter (\cref{tab:prompt_protocol}). Enlarging the template set by a factor of nearly four leaves the ordering exactly as it was, at Spearman $\rho = +1.00$, and moves the scores by $-0.009$ on average. \model ranks first under both banks and its mean paired advantage over PathGen-L/14, the strongest baseline on the aggregate, is $0.086$ under the reported bank (task-bootstrap \qty{95}{\percent} interval $0.042$--$0.136$) and $0.070$ under the six-template bank ($0.023$--$0.123$).

\begin{table}[t]
\centering
\small
\caption{The nine-dataset patch benchmark under two prompt protocols that differ only in how many ways a class is phrased. Both use the same single canonical class name per dataset. ``6/class'' is the six-template bank. ``22/class'' is the 22-template bank reported in \cref{fig:patch}, which is the set of prompt frames released with CONCH. Values are the mean chance-corrected balanced accuracy over the nine datasets, with the rank under each protocol in parentheses. $\Delta$ is the second column minus the first. Models are ordered by their score under the reported protocol.}
\label{tab:prompt_protocol}
\renewcommand{\arraystretch}{1.15}
\setlength{\tabcolsep}{5pt}
\begin{tabular}{lccr}
    \toprule
    \textbf{Model} & \textbf{6/class} & \textbf{22/class} & \textbf{$\Delta$} \\
    \midrule
    \textbf{\model}        & 0.554 (1) & \textbf{0.546} (1) & $-0.008$ \\
    PathGen-L/14           & 0.484 (2) & 0.461 (2) & $-0.023$ \\
    KEEP                   & 0.459 (3) & 0.448 (3) & $-0.011$ \\
    CONCH                  & 0.421 (4) & 0.405 (4) & $-0.016$ \\
    PathCLIP               & 0.388 (5) & 0.393 (5) & $+0.005$ \\
    QuiltNet-B/32          & 0.241 (6) & 0.240 (6) & $-0.001$ \\
    CLIP-L/14              & 0.208 (7) & 0.195 (7) & $-0.014$ \\
    \midrule
    Mean excluding \model  & & & $-0.009$ \\
    \bottomrule
\end{tabular}
\end{table}

\section{Read-out sensitivity checks} \label{sec:readout_checks}

Three checks vary a read-out choice made in the Methods while holding the model, the images and the cached embeddings fixed. None of them changes a conclusion reported in the main text.

\paragraph{Positive-class vocabulary on the melanoma cohorts.} The shared lymph-node vocabulary of \cref{ln_scoring} names carcinoma, whereas two external cohorts are melanoma. We re-scored both melanoma cohorts for every model with a histology-agnostic positive vocabulary (``metastatic tumor''/``metastatic malignancy''). \model's cohort AUROCs were unchanged at $0.997$ and $0.979$. Across the reported models the largest change in either cohort was $0.045$ and the mean absolute change was $0.008$, so the result is not an artifact of the carcinoma wording.

\paragraph{Slide aggregator.} The slide score is the maximum positive-class probability over retained tiles, and the result does not depend on the extreme-value behaviour of that maximum. Replacing it with the mean of the five highest tile probabilities changes \model's cohort-macro external AUROC by $0.003$ ($0.959$ to $0.956$), moves no single cohort's AUROC by more than $0.014$, and leaves the two slide scores correlated at Spearman $\rho = 0.97$.

\paragraph{Duplicate captions in ARCH.} \num{1083} of the \num{7576} ARCH captions are shared by more than one figure, so a model can be scored as wrong for retrieving a caption that is textually correct. Recomputing Recall@5 with identical captions collapsed to a single candidate shifted every model's Recall@5 by at most $0.02$ and preserved the ordering.

\section{The argmax decision rule on binary tasks} \label{sec:argmax_rule}

\begin{figure*}[!htbp]
    \centering
    \includegraphics[width=0.8\textwidth]{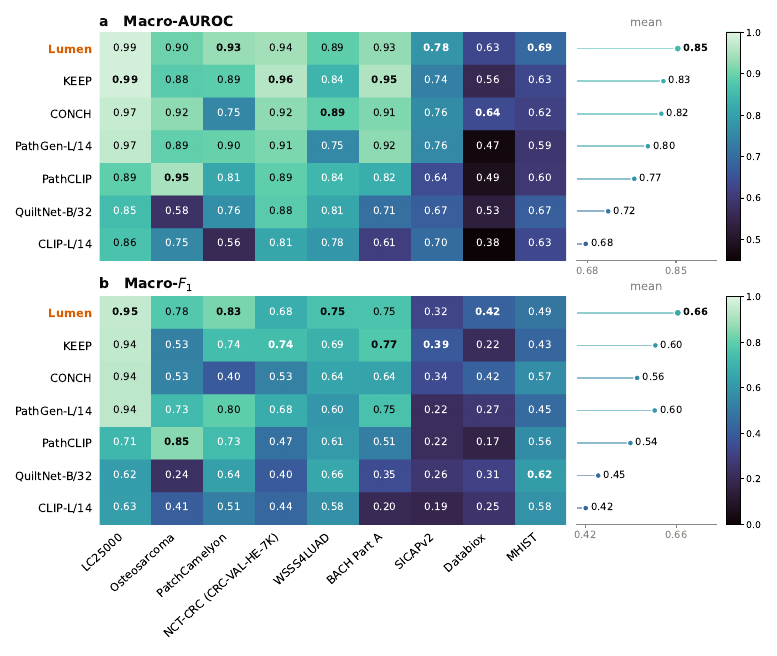}
    \caption{\textbf{Complementary zero-shot patch-benchmark metrics for all seven models on all nine datasets.} Cells contain prompt-ensemble values under the same 22-template canonical protocol as \cref{fig:patch}, so these panels are directly comparable with the main-text figure. Panels show \textbf{(a)} macro-AUROC, and \textbf{(b)} macro-$F_1$. The strip to the right of each heatmap gives each model's mean across the nine datasets. The highest mean is in bold. Unlike balanced accuracy, neither metric depends on where the argmax decision boundary falls.}
    \label{fig:supp_patch_metrics}
\end{figure*}

The read-out rule can dominate a binary result. The sharpest instance is CONCH on PatchCamelyon, where our protocol reports a balanced accuracy of $0.530$. This is a decision failure and not a representation failure. The zero-shot rule is an argmax over the two class vectors, a threshold at a margin of zero, and CONCH's margin on this dataset is centred at $-0.211$. Therefore, the cut falls in its tail and only \qty{3.0}{\percent} of patches are called positive on a set that is half positive. A linear probe on the vision embeddings reaches $0.916$ and the prompted margin still ranks at an AUROC of $0.751$. No conclusion here rests on that number. Ranked by mean macro-AUROC, \model is first at $0.853$ (Supplementary \cref{fig:supp_patch_metrics}), and dropping PatchCamelyon from the aggregate leaves \model first at $0.531$ against $0.546$ with it.

\section{Lymph-node cohorts and comparison with MetAssist~2.0} \label{sec:ln_map_atlas}

\begin{figure}[!htbp]
    \centering
    \includegraphics[width=1\linewidth]{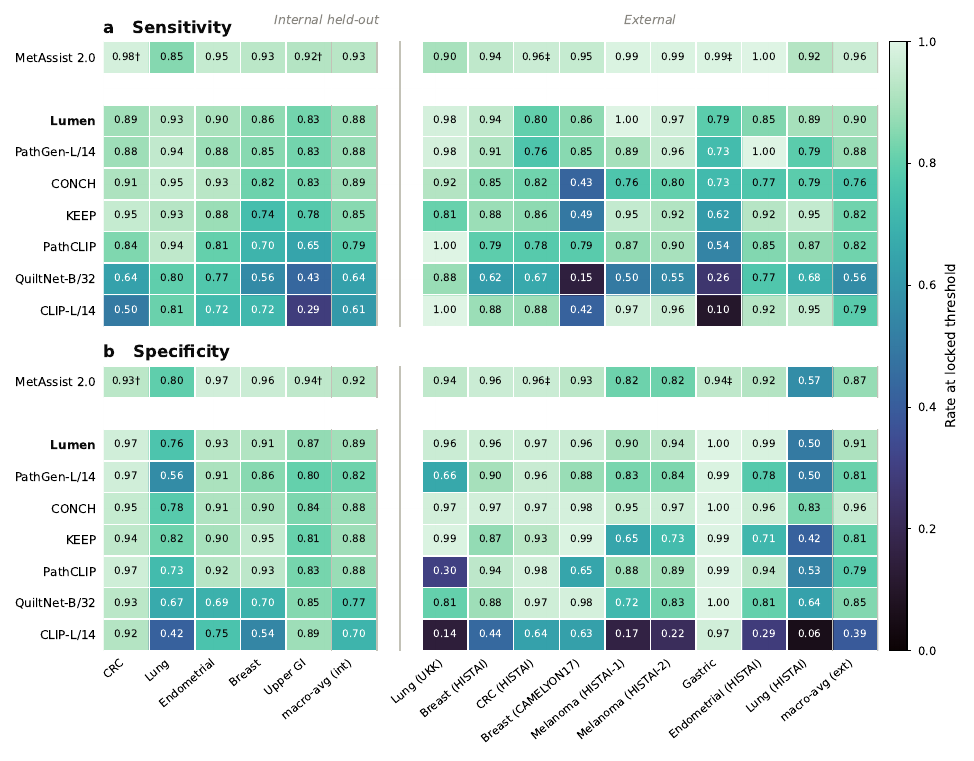}
    \caption{\textbf{Calibrated sensitivity and specificity for the lymph-node
    benchmark.} Balanced accuracy (main text, \cref{fig:ln_main}) is the mean of these two quantities and hides the split between them. \textbf{(a)} Sensitivity and
    \textbf{(b)} specificity of every model on every cohort, on a shared $0$--$1$ scale. The five internal held-out organ groups and the nine external cohorts each form a block closed by its own macro-average column, and the model rows are ordered by external balanced accuracy. The top row, set apart, is MetAssist~2.0 \cite{garcia2026metassist}, transcribed from its published table. $\dagger$ marks the two internal cohorts that contain MetAssist~2.0's own training slides. $\ddagger$ marks two external cohorts whose slides it never saw, but whose organ it was trained on.}
    \label{fig:ln_supp}
\end{figure}

\cref{tab:ln_cohorts} gives the cohorts the slide-level evaluation runs on, with patient and slide counts and the metastasis-positive and negative split of each. \cref{fig:ln_supp} then opens the headline balanced accuracy into the two quantities it averages, for every model on every cohort.

\begin{table}[t]
\centering
\caption{Lymph-node metastasis cohorts used for slide-level evaluation, with patient and slide counts and metastasis-positive/negative splits. Patient counts use the source case identifier. The public gastric cohort provides no case identifier, so each of its slides is counted individually.}
\label{tab:ln_cohorts}
\renewcommand{\arraystretch}{1.15}
\setlength{\tabcolsep}{5pt}
\begin{tabular}{llrrrr}
    \toprule
    \textbf{Cohort} & \textbf{Organ} & \textbf{Patients} & \textbf{WSIs} & \textbf{Positive} & \textbf{Negative} \\
    \midrule
    \multicolumn{6}{l}{\textit{Internal (ITMP, University of Bern)}} \\
    ITMP & Breast & 236 & 298 & 110 & 188 \\
    ITMP & Colorectal & 428 & 3,259 & 758 & 2,501 \\
    ITMP & Endometrial & 98 & 446 & 110 & 336 \\
    ITMP & Lung & 147 & 415 & 202 & 213 \\
    ITMP & Upper gastrointestinal & 166 & 800 & 250 & 550 \\
    \textit{Internal subtotal} & & 1,075 & 5,218 & 1,430 & 3,788 \\
    \quad calibration (\qty{20}{\percent} of patients) & & 215 & 1,004 & 280 & 724 \\
    \quad held-out test (\qty{80}{\percent}) & & 860 & 4,214 & 1,150 & 3,064 \\
    \midrule
    \multicolumn{6}{l}{\textit{External (public and institutional data sources, held out in full)}} \\
    CAMELYON17 & Breast & 200 & 921 & 306 & 615 \\
    HISTAI & Breast & 109 & 111 & 34 & 77 \\
    HISTAI & Colorectal & 32 & 256 & 51 & 205 \\
    HISTAI & Endometrial & 51 & 90 & 13 & 77 \\
    HISTAI & Lung & 44 & 74 & 38 & 36 \\
    HISTAI (batch 1) & Melanoma & 21 & 98 & 38 & 60 \\
    HISTAI (batch 2) & Melanoma & 35 & 164 & 71 & 93 \\
    UKK Cologne & Lung & 87 & 255 & 136 & 119 \\
    Public gastric & Gastric & (399) & 399 & 201 & 198 \\
    \textit{External subtotal} & & 978 & 2,368 & 888 & 1,480 \\
    \midrule
    \textbf{Total (evaluable)} & & \textbf{2,053} & \textbf{7,586} & \textbf{2,318} & \textbf{5,268} \\
    \bottomrule
\end{tabular}
\end{table}

Two things about MetAssist~2.0 have to be held in mind when reading that comparison, and they are what the rest of this section is about. The first is what it was trained on. MetAssist~2.0 was trained on lymph nodes from the same centre as our internal cohort and used no external training data. Its published results on the external CRC (HISTAI) and gastric cohorts are also considered in-domain because it was trained on those organs, although not on those specific slides. These four cohorts are marked separately in \cref{fig:ln_supp}, and MetAssist~2.0 performs better on all four. The remaining ten cohorts are domain-generalization cohorts for MetAssist~2.0 and are held out for both models.

\begin{figure}[!htbp]
    \centering
    \includegraphics[width=1\linewidth]{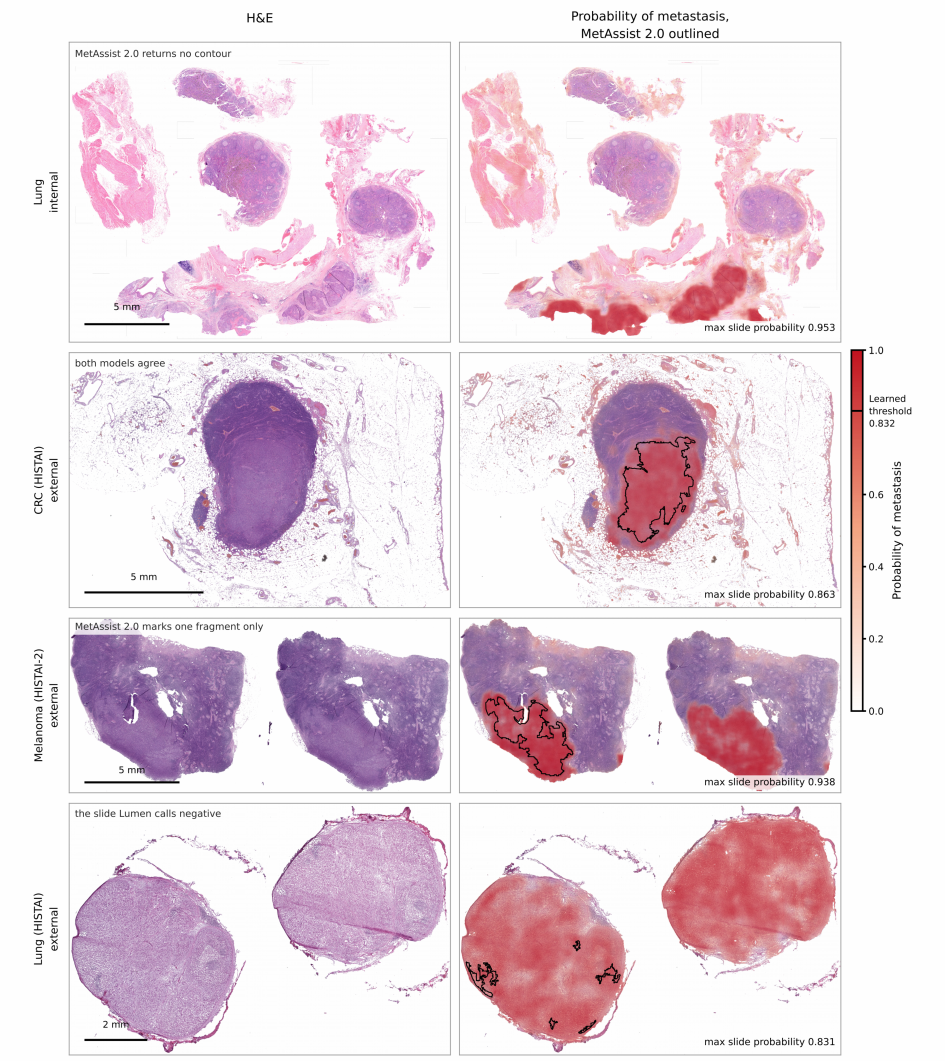}
    \caption{\textbf{Metastasis maps on four lymph-node slides.} One row per slide, the cohort named at the left. \textbf{Left column}, the H\&E image and \textbf{Right column}, the same field under \model's probability and MetAssist~2.0 outline. Each panel's max slide metastasis probability is reported at the bottom right of the right column.}
    \label{fig:ln_map_atlas}
\end{figure}

The second is how it works. By design, MetAssist~2.0 locates lymph nodes and looks for metastasis inside them, so a node it does not recognise is a node it never examines. \model{} scores tiles with no notion of a node at all. \cref{fig:ln_map_atlas} shows four slides, chosen from the fourteen cohorts to show a few special cases. The three top rows are slides \model{} calls positive, and the fourth is the one slide in the set it calls negative. That difference in design decides three of the four rows, and it does not always favour \model{}.

On the internal lung slide MetAssist~2.0 correctly returns no contour, because the tumour in frame appears to be primary site and a model that reasons about nodes is right to leave it alone. This row is a limitation of a tile-wise read-out rather than a success. What the prompt recovers is tumour, and on a slide where tumour and nodal metastasis come apart, that distinction is one \model{} does not make.

The colorectal HISTAI slide is the row where the two agree. \model's high-probability field sits on the deposit MetAssist~2.0 outlines, and neither model claims tissue the other leaves alone.

The melanoma slide carries two fragments of one node. MetAssist~2.0 marks the fragment it recognises and leaves the other unmarked, which is again a failure to identify the node rather than a failure to see the tumour in it. \model{} responds to the tumour in both fragments.

On the lung HISTAI slide MetAssist~2.0 produces no usable tumour contour, and \model{} gives the tumour a high probability that still falls $0.001$ below its locked operating point, so it is a near miss and the slide is not flagged as metastatic.

\end{document}